\documentclass[letterpaper,10pt,conference]{ieeeconf}

\IEEEoverridecommandlockouts
\usepackage{cite}
\usepackage{amsmath,amssymb,amsfonts}
\usepackage{siunitx}
\usepackage{algorithmic}
\usepackage{graphicx}
\usepackage{textcomp}
\usepackage{xcolor}
\usepackage{color, soul}
\sethlcolor{red}
\usepackage{tabularx}
\usepackage{diagbox}
\usepackage{array}
\def\BibTeX{{\rm B\kern-.05em{\sc i\kern-.025em b}\kern-.08em
    T\kern-.1667em\lower.7ex\hbox{E}\kern-.125emX}}
\begin{document}

\title{A Semantic and Occlusion-Aware GM-PHD Filter}

\author{Jovan Menezes$^{1}$ and Mark Campbell$^{1}$%
\thanks{This work was supported by NSF CPS grant CNS-2211599 and NSF FRR grant IIS-2305532. The authors acknowledge Sushrut Surve for helpful and insightful discussions.}
\thanks{$^{1}$Jovan Menezes and Mark Campbell are with the Sibley School of Mechanical and Aerospace Engineering, Cornell University, Ithaca, NY, USA.
\texttt{\{jcm483, mc288\}@cornell.edu}}
}

\maketitle
\begin{abstract}
This paper proposes a new birth model including semantic information derived from deep learning to create an occlusion-aware Gaussian Mixture Probability Hypothesis Density (GM-PHD) filter.
Unlike prior approaches that rely on simplistic or uniform assumptions, the proposed Semantic-Occlusion Aware (S-OA) birth model defines initialization terms by explicitly considering regions of occlusion and by leveraging semantic information about the environment.
This enables the filter to accurately represent where new objects are more likely to appear, thereby improving tracking performance in complex and high-density driving scenarios.
The method is evaluated through Monte Carlo simulations and experiments on the KITTI dataset.
Performance is assessed by measuring the latency between first detection and track initiation, along with the mean absolute cardinality error and the Optimal Subpattern Assignment (OSPA) metric.
Results demonstrate that the S-OA birth model reduces initialization delay in occlusion-heavy settings, matching or outperforming the strongest baseline in approximately 70$\mathbf{\%}$ of cases.
A sensitivity analysis of birth model weights is also provided.
Overall, the findings underscore the benefits of integrating occlusion reasoning and semantic priors into Bayesian tracking frameworks for autonomous driving.
\end{abstract}


\section{Introduction}
\label{Section:Introduction}
Reliable multi-target tracking (MTT) is essential for autonomous vehicles, supporting safe navigation in dynamic, multi-agent traffic environments.
However, it remains challenging due to sensor noise, occlusions, false detections, and uncertainty in both the number and states of objects \cite{b2, b4}.
MTT methods typically follow either the Tracking-by-Detection (TBD) or Joint Detection and Tracking (JDT) frameworks, and recent advances in both have incorporated deep learning.
Although these approaches achieve strong performance through large-scale data and appearance-based association, they rely heavily on extensive training data, struggle with out-of-distribution scenes, offer limited uncertainty modeling, and incur high computational costs for data association in dense, cluttered driving scenarios.

\begin{figure}[t]
    \centering
    \includegraphics[width=\linewidth]{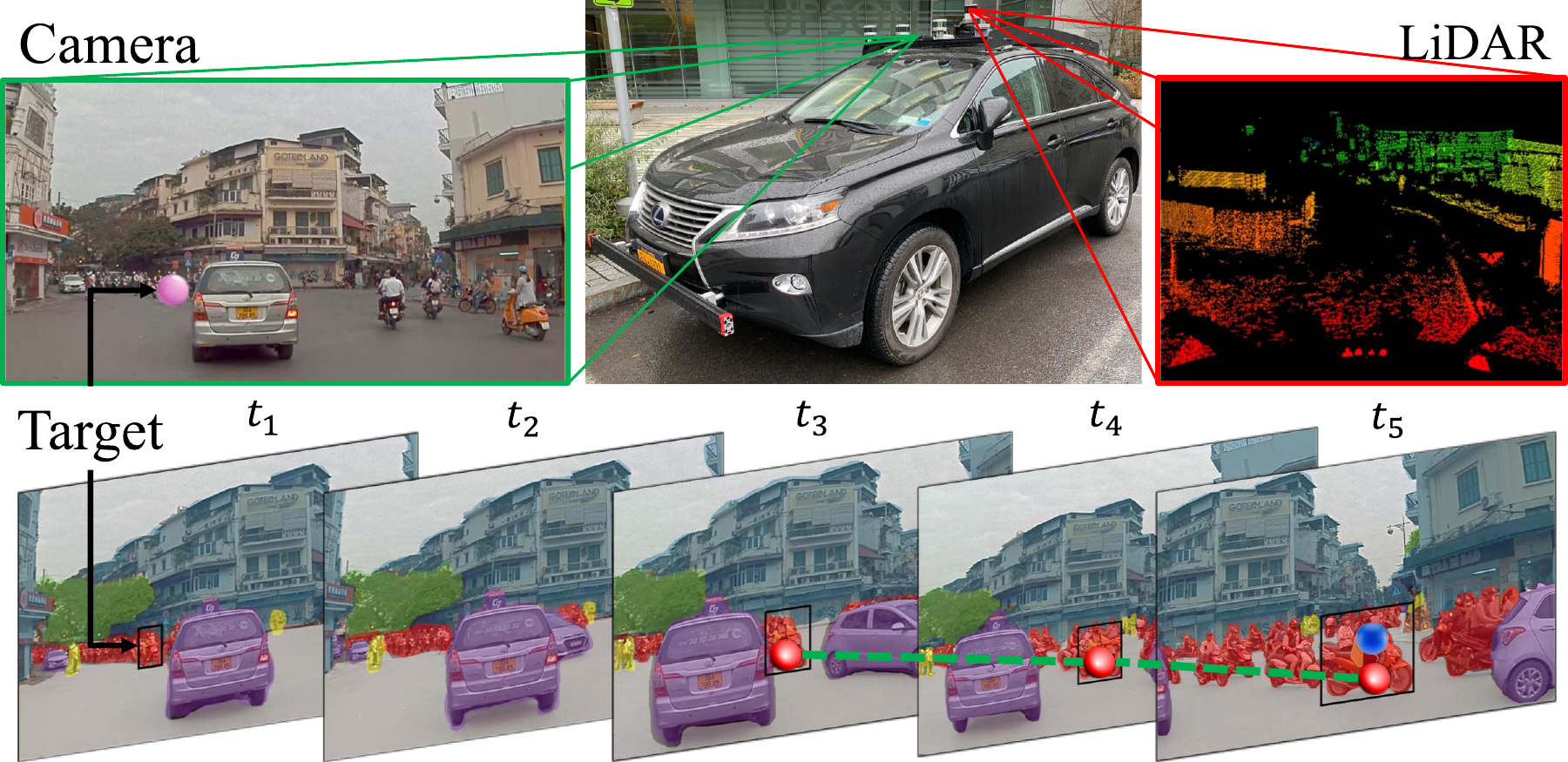}
    \caption{MTT in a dense driving scenario. A target (marked in pink in the camera frame) enters the scene at $t_1$, gets occluded at $t_2$, and reappears at $t_3$. Our S-OA GM-PHD filter leverages occlusion reasoning and scene semantics to initiate tracking (marked in red at $t_3$) more quickly than baseline methods (marked in blue at $t_5$).}
    \label{fig:01}
\end{figure}

A principled alternative to learning-based methods is offered by Bayesian probabilistic frameworks that mainly follow the TBD paradigm, such as the Joint Probabilistic Data Association (JPDA) filter \cite{b5}, Multiple Hypothesis Tracking (MHT) filter \cite{b6}, and Random Finite Set (RFS) theory \cite{b7}.
These approaches do not rely on large-scale training data and naturally quantify uncertainty.
The key distinction between these approaches lies in how they address the data association problem: JPDA averages over associations for each target, MHT maintains multiple competing hypotheses over time, and RFS models the entire multi-target state as a probabilistic set.
Among RFS-based filters, the Probability Hypothesis Density (PHD) filter \cite{b8} stands out by eliminating the need for explicit data association.
The PHD filter provides a transparent and interpretable probabilistic framework that naturally handles clutter, missed detections, and an unknown and varying number of targets.
These characteristics make it particularly suitable for MTT in autonomous vehicles, where highly dynamic environments, dense traffic, and frequent target entries and exits are prevalent, as seen in Fig. \ref{fig:01}.

The GM-PHD filter models the target intensity function, which represents the expected spatial density of targets, as a mixture of weighted Gaussian components.
The GM-PHD filter has been successfully applied across various domains \cite{b9, b10, b15}.
However, the formulation of the birth model remains a key challenge for the PHD filter and the wider class of RFS theory-based filters, as it directly governs the introduction of new targets into the scene.
Existing approaches to defining birth models are often computationally expensive, rely on restrictive assumptions, require prior map data, or necessitate careful tuning of mixture parameters to achieve acceptable tracking performance.
Standard GM-PHD filter formulations assume that potential birth locations are known a priori, and define Gaussian birth components in these regions \cite{b12}.
In autonomous driving scenarios, this assumption proves inefficient, as new targets may appear anywhere within the sensor's field of view (FOV), often emerging from behind occlusions.
Consequently, the birth model must consider the entire surveillance volume.
Frequent creation and deletion of perceived targets can also negatively affect downstream planning and control, leading to unstable or stop-and-go driving behavior.

A simple alternative is to uniformly distribute birth terms across the workspace, but this is computationally inefficient and impractical for real-time use.
A hybrid approach combining a uniform distribution with selected Gaussian components is proposed in \cite{b13}; however, its computational cost grows significantly with workspace size.
Adaptive methods in \cite{b14, b16} update the birth model at each scan using incoming measurements, but they are highly sensitive to clutter and prone to tracking false positives, an important limitation in safety-critical applications such as autonomous driving.

Most existing methods do not explicitly incorporate scene context, occlusions, or semantic structure into the birth model.
Although \cite{b17, b19} integrate semantic information into the PHD filter, they focus mainly on target classification rather than broader environmental understanding.
In autonomous driving, new targets often emerge from occluded areas, such as behind vehicles or infrastructure, and their appearance depends strongly on scene semantics, for example, pedestrians are more likely near crosswalks.
Incorporating occlusion awareness and semantic priors into the birth process can therefore improve initialization, reduce false positives, and enhance overall tracking performance.

Motivated by the need for improved occlusion reasoning we introduce a novel birth model for the GM-PHD filter that explicitly integrates both occlusion information and environmental semantics to accurately represent the spatial distribution of new targets.
By leveraging these contextual cues, the S-OA GM-PHD filter shortens the latency in initiating tracks for newly born targets and improves robustness in complex, cluttered scenes.
We present the following contributions of this work:
\begin{itemize}
\item A novel birth model for the GM-PHD filter that explicitly incorporates scene semantics and occlusion information, enabling more accurate target initialization in dense and dynamic environments.
\item A comprehensive evaluation of the S-OA birth model through extensive Monte Carlo simulations and experiments on the KITTI benchmark dataset, with performance compared against multiple established baselines.
\item An analysis of tracking robustness of the S-OA birth model relative to the uniform and partially uniform birth model under varying model parameters, addressing a gap that has received limited attention in previous studies.
\end{itemize}

\section{Background}
\label{Section:Background}
The formulation of the PHD filter follows references \cite{b8, b12, b18}.
At time step $k$, we consider a set $\mathbf{X}_{k}$ of $n_k$ targets in the state space $\mathcal{X} \in \mathbb{R}^{n_x}$, where each target state is denoted by $\mathbf{x}_{1,k}, \dots, \mathbf{x}_{{n_k},k}$.
Similarly, the sensor receives a set $\mathbf{Z}_{k}$ of $m_k$ measurements represented by $\mathbf{z}_{1,k}, \dots, \mathbf{z}_{{m_k},k}$ from the observation space $\mathcal{Z} \in \mathbb{R}^{n_z}$.
RFS models the multi-target state $\mathbf{X}_{k}$ and multi-target measurement $\mathbf{Z}_{k}$ collections as finite sets, with no inherent ordering among their elements.
Every time step, certain targets may disappear, while others survive and evolve into new states, and additional targets may be introduced into the scene.
Owing to imperfections in the sensor and the detection process, some surviving or newly born targets may remain undetected, while the observation set $\mathbf{Z}_{k}$ may also contain false positives.

Given a multi-target state $\mathbf{X}_{k-1}$, the multi-target state $\mathbf{X}_{k}$ is expressed as the union of surviving targets, spawned targets, and spontaneous births:
\begin{equation}
\mathbf{X}_{k} = \left[ \bigcup_{\zeta \in \mathbf{X}_{k-1}} S_{k|k-1}(\zeta) \right] \cup  \left[ \bigcup_{\zeta \in \mathbf{X}_{k-1}} B_{k|k-1}(\zeta) \right] \cup \Gamma_k
\label{eq:1}
\end{equation}
\noindent where $S_{k|k-1}(\zeta)$ is the RFS of surviving targets originating from a target with previous state $\zeta$, $\Gamma_k$ is the RFS of spontaneously born targets, and $B_{k|k-1}(\zeta)$ is the RFS of targets spawned from a target with previous state $\zeta$.
Given a sequence of measurements up to $k$, the objective of the Bayesian multi-target recursive filter is to estimate the multi-target posterior density $p_{k}(\mathbf{X}_k|\mathbf{Z}_{1:k})$.
However, the direct computation of this posterior is intractable due to its exponential complexity.
To address this challenge, \cite{b8} introduced an approach that propagates only the first-order statistical moment of $p_{k}(\mathbf{X}_k|\mathbf{Z}_{1:k})$, known as the intensity function or the PHD.
Formally, the PHD of targets in the state space $\mathcal{X}$ is defined as an intensity function $\nu(x_k)$, such that its integral over a region $\mathcal{S} \subseteq \mathcal{X}$ yields the expected number of targets within $\mathcal{S}$, \textit{i.e.}, $N(\mathcal{S}) = \int_{\mathcal{S}}\nu(x_k) dx_k$.

Analogous to single-target Bayesian filters, the PHD filter estimates the posterior intensity $\nu_{k}(x_k)$ through a two-step recursion: prediction and update.
In the prediction step, the predicted PHD $\nu_{k|k-1}(x_k)$ is obtained from the posterior PHD from the previous time step, $\nu_{k-1}(\zeta)$, using all measurements up to time $k-1$.
This corresponds to a one-step prediction:
\begin{align}
    \nu_{k|k-1}(x_k) = &\int (p_{S,k}(\zeta)~f_{k|k-1}(x_k|\zeta)~\nu_{k-1}(\zeta))d\zeta + \nonumber \\
    &\int(\beta_{k|k-1}(x_k|\zeta)~\nu_{k-1}(\zeta)) d\zeta + \gamma_{k}(x_k)
    \label{eq:2}
\end{align}
\noindent where $p_{S,k}(\zeta)$ is the probability that a target with previous state $\zeta$ survives to time $k$, $f_{k|k-1}(x_k|\zeta)$ is the state transition density from $\zeta$ to $x_k$, $\beta_{k|k-1}(x_k|\zeta)$ is the intensity of the RFS $B_{k|k-1}(\zeta)$ of a target spawned with previous state $\zeta$, and $\gamma_{k}(x_k)$ is the intensity of the spontaneous birth RFS $\Gamma_k$.
In the update step, the posterior intensity $\nu_{k}(x_k)$ is obtained by incorporating the measurements $\mathbf{Z}_k$ with $\nu_{k|k-1}(x_k)$:
\begin{align}
    &\nu_{k}(x_k) = [1-p_{D,k}(x_k)]~\nu_{k|k-1}(x_k) \label{eq:3} \\
    &+ \sum_{z_k \in \mathbf{Z}_k} \frac{p_{D,k}(x_k)~g_k(z_{k}|x_{k})~\nu_{k|k-1}(x_k)}{\kappa_k(z_{k}) + \int (p_{D,k}(x_k)~g_k(z_k|x_k)~\nu_{k|k-1}(x_k))dx_k} \nonumber
\end{align}
\noindent where $p_{D,k}(x_k)$ is the probability of detection for a target in state $x_k$, $g_k(z_{k}|x_{k})$ is the sensor likelihood function, and $\kappa_k(z_{k})$ is the intensity of the clutter RFS.

Several formulations of the PHD filter have been proposed in literature (\textit{e.g.}, \cite{b12}, \cite{b18}).
In this work, we adopt the GM-PHD filter, in which the PHDs $\nu_{k|k-1}(x_k)$ from (\ref{eq:2}) and $\nu_{k}(x_k)$ from (\ref{eq:3}) are approximated using a weighted sum of Gaussian components in a mixture:
\begin{equation}
    \nu_{k|k-1}(x_{k}) = \sum_{i=1}^{J_{k|k-1}} {w_{k|k-1}^{(i)}~\mathcal{N}(x;m_{k|k-1}^{(i)},P_{k|k-1}^{(i)})}
    \label{eq:4}
\end{equation}
\begin{equation}
    \nu_{k}(x_{k}) = \sum_{i=1}^{J_{k|k}} {w_{k|k}^{(i)}~\mathcal{N}(x;m_{k|k}^{(i)},P_{k|k}^{(i)})}
    \label{eq:5}
\end{equation}
\noindent where $J_{.|.}$ denotes the number of Gaussian components, $w_{.|.}^{(i)}$ is the weight of the $i$-th component, $m_{.|.}^{(i)}$ its mean, and $P_{.|.}^{(i)}$ its covariance.
The predicted intensity in (\ref{eq:4}), analogous to the multi-target state $\mathbf{X}_k$ defined in (\ref{eq:1}), is composed of three components: the intensity of surviving targets $\nu_{S,k|k-1}(x_{k})$, the intensity of targets spawned from previously existing targets $\nu_{\beta,k|k-1}(x_{k})$, and the intensity of spontaneously birthed targets $\gamma_{k}(x_k)$.
Each of these components is represented as a Gaussian mixture, yielding:
\begin{equation}
    \nu_{k|k-1}(x_{k}) = \nu_{S,k|k-1}(x_{k}) + \nu_{\beta,k|k-1}(x_{k}) + \gamma_{k}(x_k)
    \label{eq:6}
\end{equation}
Various birth models have been proposed in literature.
However, most existing approaches fail to account for occlusions or leverage semantic information from the environment, while also being computationally demanding.
In the following section, we present our birth model formulation, which explicitly addresses these limitations.

\section{Semantic-Occlusion Aware GM-PHD Filter}
\label{Section:Method}
\begin{figure}[t]
    \centering
    \includegraphics[width=0.95\linewidth]{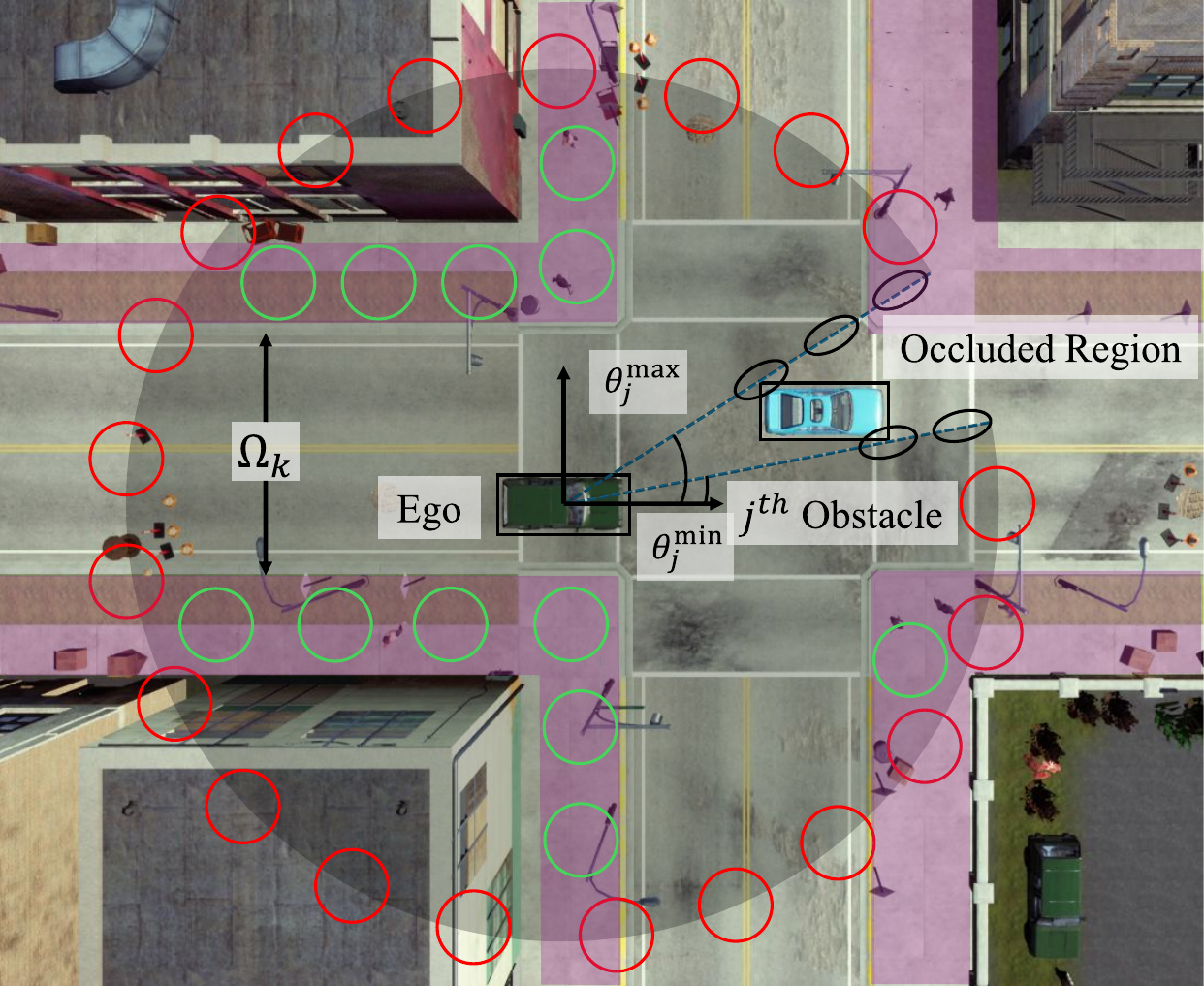}
    \caption{Visualization of the GM components in the S-OA birth model. Black, green, and red components correspond to the occlusion-based terms (\ref{eq:10}), semantic-based terms (\ref{eq:13}), and boundary terms at the edge of the sensor's FOV (\ref{eq:15}), respectively.}
    \label{fig:02}
\end{figure}

We adopt the GM-PHD filter as it avoids the computationally expensive data association step, making it particularly suitable for real-time autonomous driving applications.
Although the S-OA birth model is demonstrated within the GM-PHD framework, the formulation is general and can be extended to other PHD filter variants or to other filters based on RFS theory, such as Bernoulli filters.
The input to our framework is a raw point cloud $\mathbf{P}_k = \{\, p_{i,k} \,\}_{i=1}^{c_k}, \quad p_{i,k} \in \mathbb{R}^3,$ where $c_k$ denotes the total number of points $p_{i,k}$ in the cloud.
The point cloud can be generated either from LiDAR sensors or via camera images in combination with deep stereo matching approaches, such as Pseudo-LiDAR++ \cite{b23}.
In the PHD recursion, the birth intensity function $\gamma_k(x_k)$ is defined over the single-target state space $x \in \mathbb{R}^{n_x}$, and in the GM implementation it is represented as:
\begin{equation}
    \gamma_k(x_k) = \sum_{i=1}^{J_{\gamma,k}} w_{\gamma,k}^{(i)} \, ~\mathcal{N}\!\left(x; ~m_{\gamma,k}^{(i)}, ~P_{\gamma,k}^{(i)}\right)
    \label{eq:7}
\end{equation}
where $J_{\gamma,k}$ is the total number of Gaussian birth components.
In our formulation, the birth intensity in (\ref{eq:7}) is defined adaptively by integrating geometric occlusion reasoning and semantic priors from the environment, thereby enabling the GM-PHD filter to hypothesize new targets in a context-aware and data-driven manner.
Fig.~\ref{fig:02} illustrates the proposed semantics- and occlusion-aware (S-OA) birth model, comprising three components: occlusion-based terms (black), semantic region terms (green), and sensor FOV terms (red).
The subsequent sections detail the construction of these components: Section~\ref{SubSection:Occlusion Birth Intensity} describes the occlusion-based birth terms, Section~\ref{SubSection:Semantic Birth Intensity} introduces the semantics-based birth terms, and Section~\ref{SubSection:FOV Boundary Birth Intensity} presents the FOV boundary-based birth terms.
Finally, Section~\ref{SubSection:Combined Birth Intensity} explains how these three components are integrated to form the complete birth model.

\subsection{Occlusion Birth Intensity}
\label{SubSection:Occlusion Birth Intensity}
The point cloud $\mathbf{P}_k$ is processed by an object detector, which provides the finite set of measurements $ \mathbf{Z}_k = \{\, \mathbf{z}_l \,\}_{l=1}^{m_k},$ with each $\mathbf{z}_l \in \mathbb{R}^{n_z}$ describing a detected object hypothesis (\textit{e.g.}, bounding box).
From $\mathbf{Z}_k$, we extract a set of obstacles $ \mathbf{O}_k = \{\, \mathbf{o}_j \,\}_{j=1}^{n_{o,k}},$ $\mathbf{o}_j \in \mathbb{R}^{n_z}$, where $n_{o,k}$ is the number of obstacles in the sensor FOV that occlude parts of the scene relative to the sensor or ego.
The sensor or ego state is expressed in the bird's-eye view (BEV) as $\mathbf{x}_{e,k} = (x_{e,k}, y_{e,k}, \theta_{e,k})$, where $(x_{e,k}, y_{e,k})$ denotes the sensor’s position and $\theta_{e,k}$ its orientation.
Each obstacle $o_j \in \mathbf{O}_k$ is parameterized in polar coordinates with respect to $\mathbf{x}_e$ as $ o_j = (r'_j, \theta'_j),$ where $r'_j \in \mathbb{R}_{+}$ is the radial distance to the obstacle centroid and $\theta'_j \in (-\pi, \pi]$ is the azimuth angle.

For each obstacle $o_j$, we define its occlusion cone $\mathcal{C}_j$ as the angular sector spanned by two rays originating at $(x_e, y_e)$ and tangent to the boundary of the obstacle.
This set describes the unobservable region behind the obstacle.
Mathematically, if $\theta_j^{\min}$ and $\theta_j^{\max}$ denote the tangent boundary angles for the $j$-th obstacle, then:
\begin{equation}
    \mathcal{C}_j = \left\{ (r_c, \theta_c) \in \mathbb{R}_+ \times (-\pi,\pi] | r_c > r'_j, \theta_j^{\min} \leq \theta_c \leq \theta_j^{\max} \right\}
    \label{eq:8}
\end{equation}

We define new Gaussian birth components along the extended occlusion boundaries $\theta_j^{\min}$ and $\theta_j^{\max}$.
Specifically, for each boundary ray $r^b_j \in \{r_j^{\min}, r_j^{\max}\}$ , we place components at radial distances:
\begin{equation}
    r \in \{ r^b_j + \Delta r, \; r^b_j + 2\Delta r, \ldots, r^b_j + L \Delta r \}
    \label{eq:9}
\end{equation}
where $\Delta r$ is the radial spacing and $L$ is the maximum number of layers behind the obstacle.
The birth intensity for the occluded regions is then a GM given by:
\begin{equation}
    \gamma_{\text{occl}, k}(x_k) = \sum_{i=1}^{J_{\text{occl},k}} w_{\text{occl},k}^{(i)} \, ~\mathcal{N}\!\left(x; ~m_{\text{occl},k}^{(i)}, ~P_{\text{occl},k}^{(i)}\right)
    \label{eq:10}
\end{equation}
where $J_{\text{occl},k}$ is the number of occlusion-based birth components.
The mean of each Gaussian component is defined as: $m_{\text{occl},k}^{(i)} = {\begin{bmatrix} r^{(i)} \cos \theta^{(i)},~r^{(i)} \sin \theta^{(i)} \end{bmatrix}}^T$ for $\theta \in \{\theta_j^{\min}, \theta_j^{\max}\}$.
The covariance is defined as anisotropic to capture increased uncertainty along the depth (radial) direction:
\begin{equation}
    P_{\text{occl},k}^{(i)} = R(\theta) \begin{bmatrix} \sigma_\parallel^2 & 0 \\ 0 & \sigma_\perp^2 \end{bmatrix} R(\theta)^\top
    \label{eq:11}
\end{equation}
where $R(\theta)$ is the 2D rotation matrix aligning the principal axis with the boundary direction, $\sigma_\parallel^2$ and $\sigma_\perp^2$ represent the variance along the depth (radial) and orthogonal directions.
Next, we integrate birth terms informed by scene semantics into the model to augment the occlusion birth intensity described in (\ref{eq:10}).

\subsection{Semantic Birth Intensity}
\label{SubSection:Semantic Birth Intensity}
At each time step, we perform semantic segmentation on $\mathbf{P}_k$, assigning each point $p_{i,k}$ a class label $l(p_{i,k}) \in \mathcal{L}$, where $\mathcal{L}$ is the set of semantic classes (\textit{e.g.}, road, sidewalk, car, etc.).
Let $\mathbf{R}_{b,k} \subseteq \mathbf{P}_k$ denote the subset of points with label corresponding to birth-relevant regions, such as pedestrian sidewalks or road entry points.
Mathematically we define $\mathbf{R}_{b,k}$ as:
\begin{equation}
    \mathbf{R}_{b,k} = \{ p_{i,k} \in  \mathbf{P}_k \;|\; l(p_{i,k}) \in \mathcal{L}_{\text{birth}} \}
    \label{eq:12}
\end{equation}
where $\mathcal{L}_{\text{birth}} \subseteq \mathcal{L}$ is the subset of semantic classes from which new targets can reasonably emerge.
We define the region of interest for birth modeling as $ \Omega_k \subseteq \mathbb{R}^2, $ corresponding to the projection of the birth-relevant point set $\mathbf{R}_{b,k}$ onto the ground plane.

The semantics-based birth intensity is modeled as a Gaussian mixture of the form:
\begin{equation}
\gamma_{\text{sem}, k}(x_k) = \sum_{i=1}^{J_{\text{sem},k}}
w_{\text{sem},k}^{(i)}~\mathcal{N}\left(x;~m_{\text{sem},k}^{(i)}, ~P_{\text{sem},k}^{(i)}\right)
\label{eq:13}
\end{equation}
where $J_{\text{sem},k}$ is the number of birth components defined based on scene semantics.
The placement of the means $m_{\text{sem},k}^{(i)}$ within $\Omega_k$ can be determined by using either uniform sampling ($m_{\text{sem},k}^{(i)} \sim \mathcal{U}(\Omega_k)$), or Density-Based Spatial Clustering (DBSCAN) algorithms, or Poisson disk sampling according to:
\begin{equation}
m_{\text{sem},k}^{(i)} \in \Omega_k \quad \text{s.t.} \quad
\| m_{\text{sem},k}^{(i)} - m_{\text{sem},k}^{(j)} \| \geq d_{\min},~\forall i \neq j
\label{eq:14}
\end{equation}
where $d_{\min}$ is the minimum Euclidean distance between any two means.
The covariances are defined as $P_{\text{sem},k}^{(i)} = \sigma_{\text{sem}}^2 \mathbf{I}$, where $\sigma_{\text{sem}}$ denotes the standard deviation of the Gaussian birth terms associated with the semantics and $\mathbf{I}$ is the identity matrix in $\mathbb{R}^{n_x}$.
This standard deviation is either set using prior knowledge or adaptively scaled according to the region’s extent or the dispersion of the cluster.

\subsection{FOV Boundary Birth Intensity}
\label{SubSection:FOV Boundary Birth Intensity}
The birth intensities defined from occlusions (\ref{eq:10}) and semantic regions (\ref{eq:13}) primarily account for targets appearing within the interior of the sensor’s FOV, such as pedestrians emerging from behind vehicles or exiting buildings.
To additionally capture targets that enter the scene directly at the periphery of the FOV, we define the boundary-based birth intensity expressed as:
\begin{equation}
\gamma_{\text{fov}, k}(x_k) = \sum_{i=1}^{J_{\text{fov},k}}
w_{\text{fov},k}^{(i)}~\mathcal{N}\left(x; m_{\text{fov},k}^{(i)}, P_{\text{fov},k}^{(i)}\right)
\label{eq:15}
\end{equation}
where $J_{\text{fov},k}$ is the number of birth components defined along the sensor FOV.
In conventional GM-PHD filter formulations, such boundary-based terms constitute the sole mechanism for initiating new tracks.
While this approach allows for detection of targets entering from outside the sensor range, it neglects objects that may emerge from within the FOV due to occlusions, thereby limiting robustness in dense or dynamic environments.
The mean locations $m_{\text{fov},k}^{(i)}$ are adaptively positioned along the FOV boundary, sampled at equidistant intervals determined from the current sensor point cloud $\mathbf{P}_k$. The covariances are modeled as isotropic, $ P_{\text{fov},k}^{(i)} = \sigma_{\text{fov}}^2 \mathbf{I},$ with variance $\sigma_{\text{fov}}^2$ chosen a priori according to the application.

\subsection{Combined Birth Intensity}
\label{SubSection:Combined Birth Intensity}
The weights $w_{\text{occl},k}^{(i)}$, $w_{\text{sem},k}^{(i)}$, and $w_{\text{fov},k}^{(i)}$ denote the expected contribution of new targets associated with their respective Gaussian birth components.
The total birth weight of the GM-PHD filter is therefore:
\begin{equation}
\sum_{i=1}^{J_{\gamma,k}} w_{\gamma,k}^{(i)} =
\sum_{i=1}^{J_{\text{occl},k}} w_{\text{occl},k}^{(i)} +
\sum_{i=1}^{J_{\text{sem},k}} w_{\text{sem},k}^{(i)} +
\sum_{i=1}^{J_{\text{fov},k}} w_{\text{fov},k}^{(i)}
\label{eq:16}
\end{equation}
This corresponds to the expected number of new targets entering the scene at time $k$, denoted by $\hat{N}_k$.
This value may be specified based on prior knowledge of the environment (\textit{e.g.}, higher in dense urban traffic) or estimated empirically from data.
A simple approach is to assign equal weights to all birth components: $w_{\gamma,k}^{(i)} = \hat{N}_k/J_{\gamma,k}$.
To account for differing reliability among birth sources, we can introduce source confidence coefficients $\rho_{s,k} \geq 0$, $s \in \{{\text{occl}, \text{sem}, \text{fov}}\}$ that represent the relative importance assigned to each birth mechanism.
The weight for the $i$-th component is:
$w_{\gamma,k}^{(i)} = (\hat{N}_k*\rho{_{s,k}^i})/ {\sum_{j=1}^{J_{\gamma,k}} \rho{_{s,k}^j}}$.
Thus, components from more reliable sources receive proportionally larger weights while preserving the total expected number of births.
The combined birth intensity is obtained as the sum of the occlusion-driven, semantic-driven, and FOV-boundary birth intensities:
\begin{equation}
    \gamma_k(x_k) = \gamma_{\text{occl}, k}(x_k) + \gamma_{\text{sem}, k}(x_k) + \gamma_{\text{fov}, k}(x_k)
    \label{eq:17}
\end{equation}

\section{Simulations}
\label{Section:Simulations}
\subsection{Simulation Setup}
\label{SubSection:Simulation Setup}
\begin{figure}[t]
    \centering
    \includegraphics[width=\linewidth]{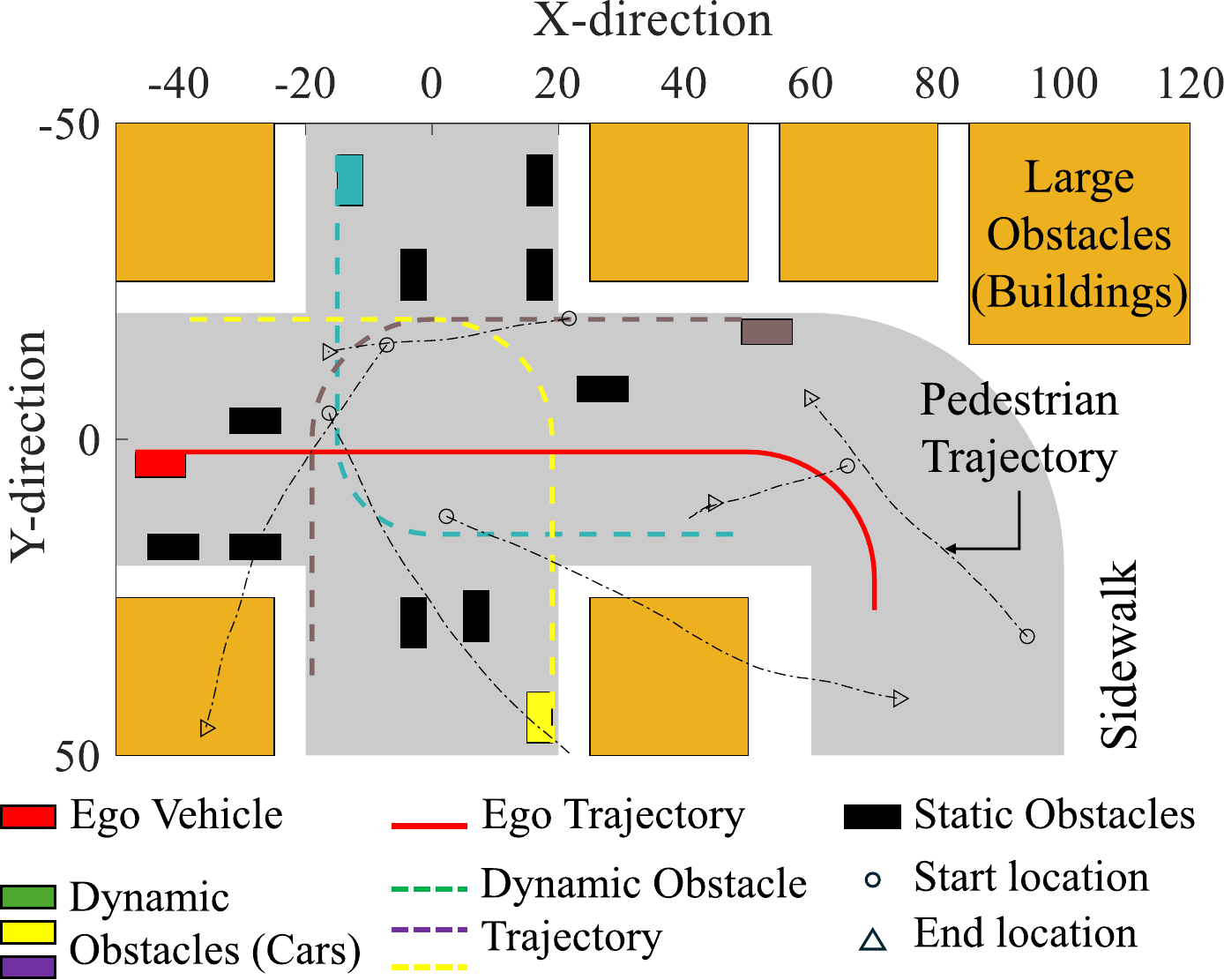}
    \caption{Simulation setup illustrating a driving scenario with multiple targets and obstacles in the environment.}
    \label{fig:03}
\end{figure}

We evaluate the proposed S-OA GM-PHD filter in a simulated environment.
For simplicity, the spawning terms in (\ref{eq:1}), (\ref{eq:2}), and (\ref{eq:6}) are omitted, without loss of generality.
The simulation consists of a 2D bird’s-eye-view intersection scenario (Fig. \ref{fig:03}).
The ego vehicle (red), equipped with an onboard sensor, starts on the left and follows a predefined trajectory.
The scene includes both dynamic and static obstacles: three moving vehicles (shown in different colors), small static objects such as parked cars (black), and large static structures such as buildings (yellow), which create extended occlusions.
The tracking task involves six pedestrians with state vector $\mathbf{x}_{i,k} = (x_{i,k},\, y_{i,k},\, v_{i,k},\, \theta_{i,k})$, where $(x_{i,k}, y_{i,k})$ denotes 2D position, $v_{i,k}$ velocity, and $\theta_{i,k}$ heading direction.
The state space is defined by $x_{i,k} \in [-50, 120]~\text{m}$, $y_{i,k} \in [-50, 50]~\text{m}$, $v_{i,k} \in [-1.0, 1.0]~\text{m/s}$, and $\theta_{i,k} \in [-\pi, \pi]$.
Pedestrians move randomly, may enter or exit buildings, and are frequently occluded by smaller obstacles.
A pedestrian's initial position is marked by ($\circ$), final position by ($\triangle$), and trajectory is shown by black dotted lines.


The sensor provides measurements of the form $\mathbf{z}_{j,k} = (r'_{j,k},\, \theta'_{j,k}),$ where $r'_{j,k}$ is the range of the $j$-th detection and $\theta'_{j,k}$ its bearing.
Each measurement may originate from a true target or from clutter, leading to false positives.
We do not estimate the ego vehicle state $\mathbf{x}_{e,k}$ and assume it to be deterministic.
To model sensor uncertainty, Gaussian zero-mean white noise is added with standard deviations of $1~\text{m}$ for range and $5^{\circ}$ for bearing.
Process noise is simulated by adding Gaussian zero-mean perturbations with standard deviations of $0.1~\text{m/s}$ for heading velocity and $3^{\circ}$ for heading angle.
In our simulations, we assume that semantic information is available.
Since the objective is to track pedestrians, the region $\Omega_k$ is defined as sidewalks, represented by the set $ \Omega^{\text{sidewalk}}_k,$
corresponding to spatial areas (illustrated in white) located between roads (depicted in grey) and buildings in Fig. \ref{fig:03}.
Birth components associated with semantic information, defined in (\ref{eq:13}) are therefore restricted to $\Omega^{\text{sidewalk}}_k$.
Since the simulated scenario contains six targets throughout the experiment, we set the expected number of newly born targets to $\hat{N}_k = 6/100$.
We set $p_{S,k}=0.99$ and $p_{D,k}=0.98$, with merging and pruning thresholds of $4$ and $1\times10^{-5}$, respectively, following \cite{b12}.

\begin{figure*}[t]
    \centering
    \includegraphics[width=0.9\linewidth]{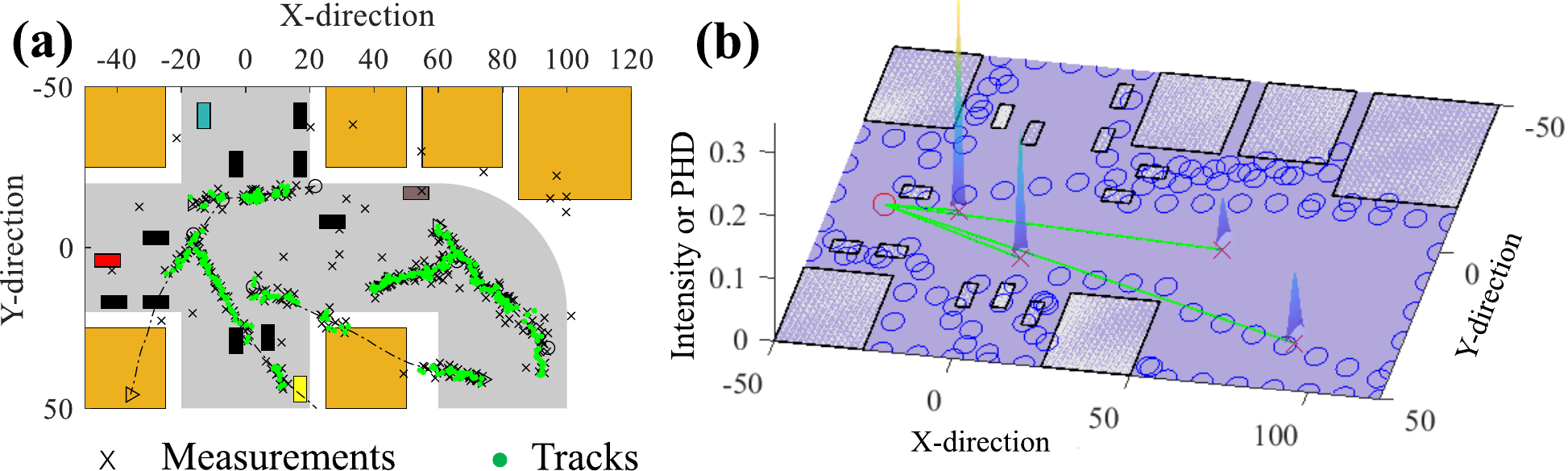}
    \caption{MTT using the GM-PHD filter with the S-OA birth model. \textbf{(a)} Simulation environment showing measurements and tracks from the filter. \textbf{(b)} 3D visualization showing the GM components in the S-OA birth model, tracked targets (green), and the intensity/PHD at a specific time step in a simulation trial.}
    \label{fig:04}
\end{figure*}

\begin{figure*}
    \centering
    \includegraphics[width=0.24\linewidth]{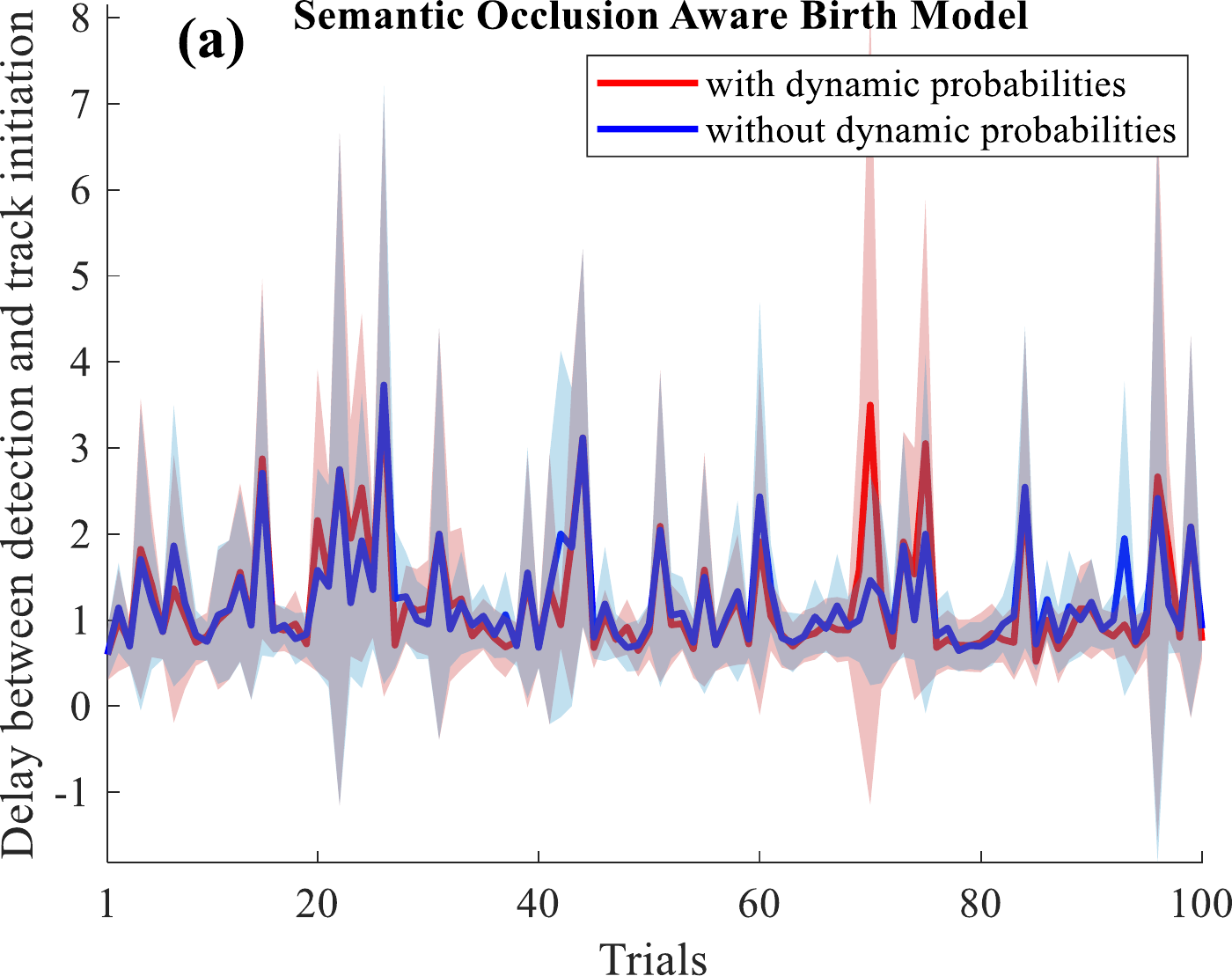}
    \vspace{0.5em}
    \includegraphics[width=0.24\linewidth]{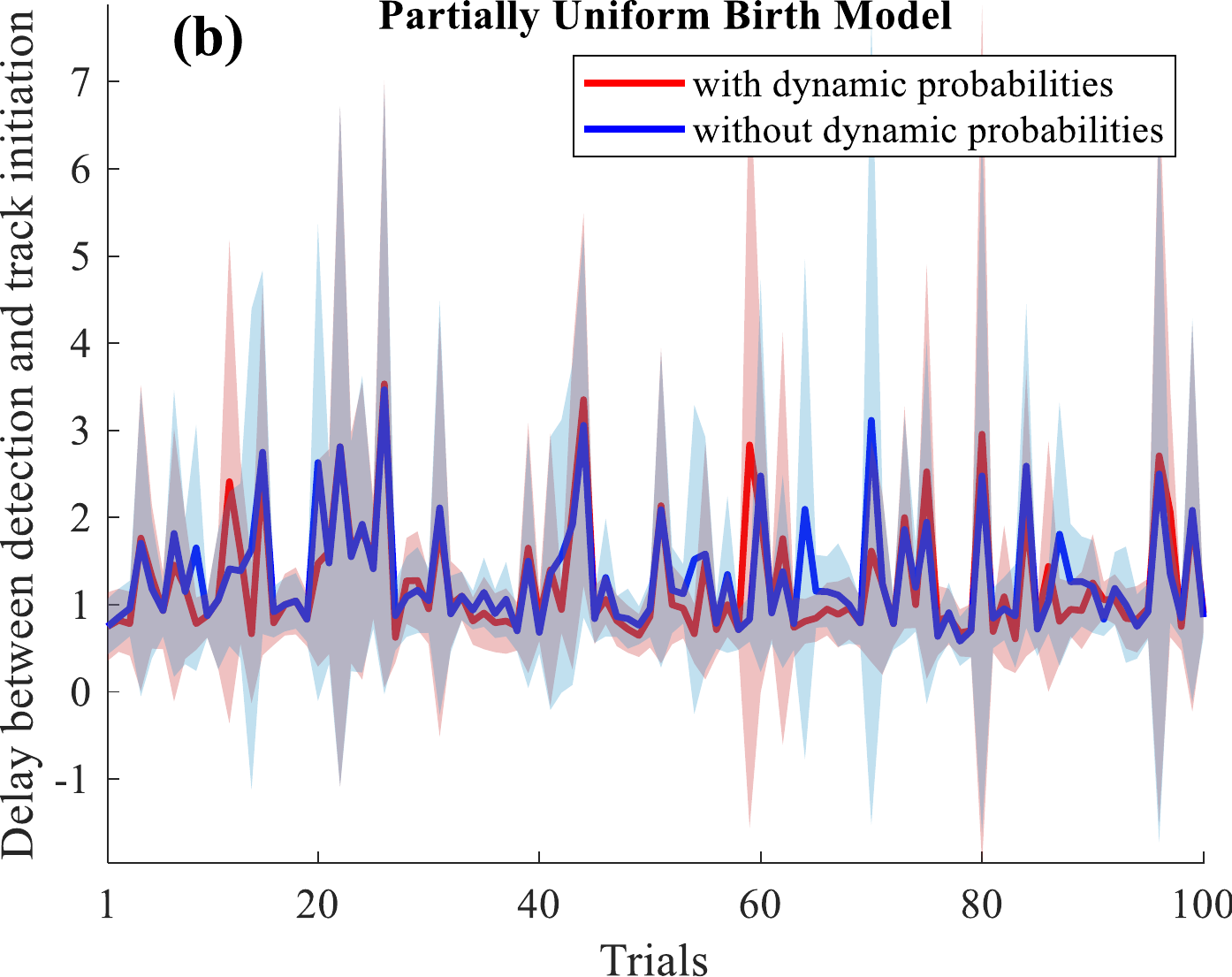}
    \includegraphics[width=0.24\linewidth]{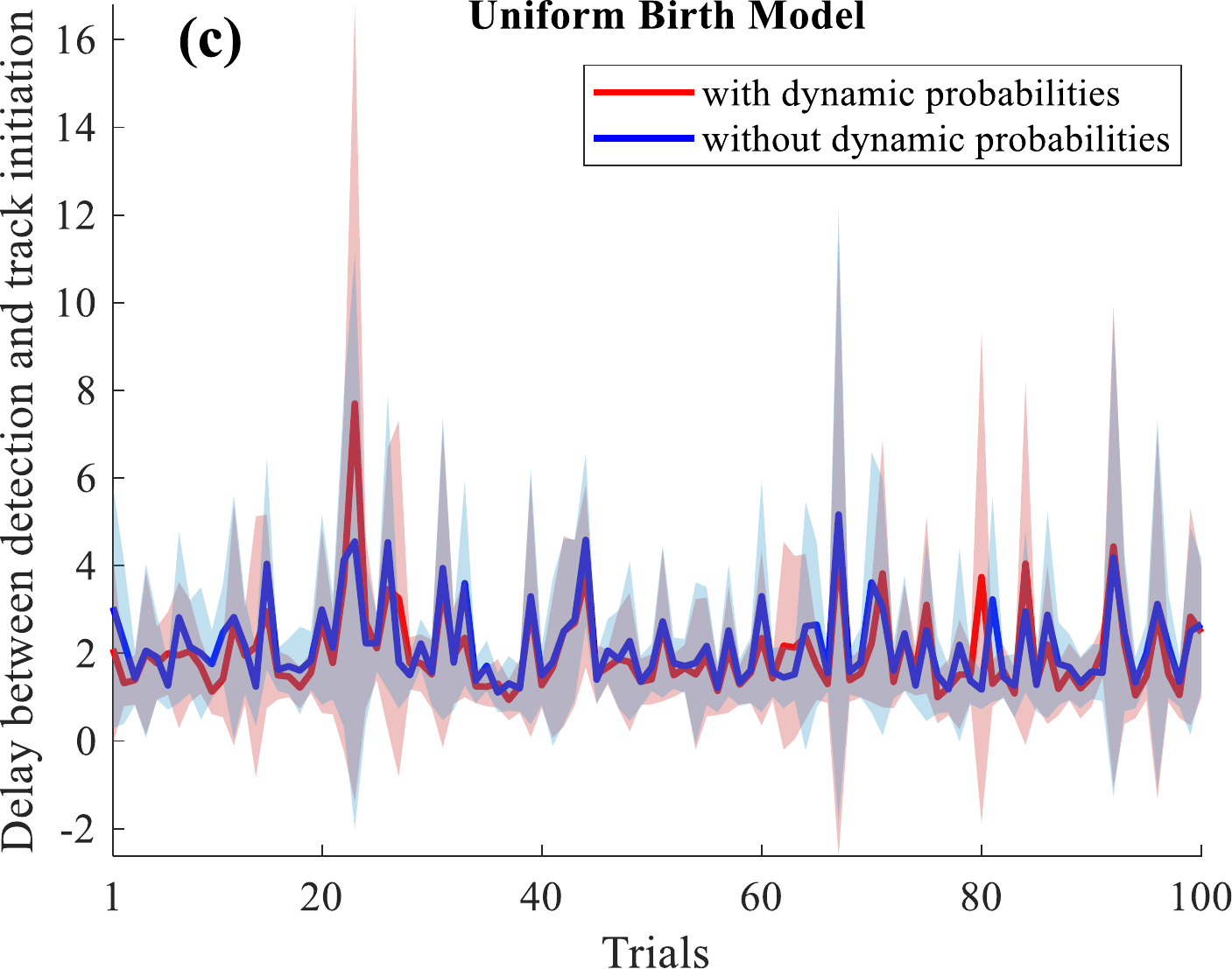}
    \includegraphics[width=0.24\linewidth]{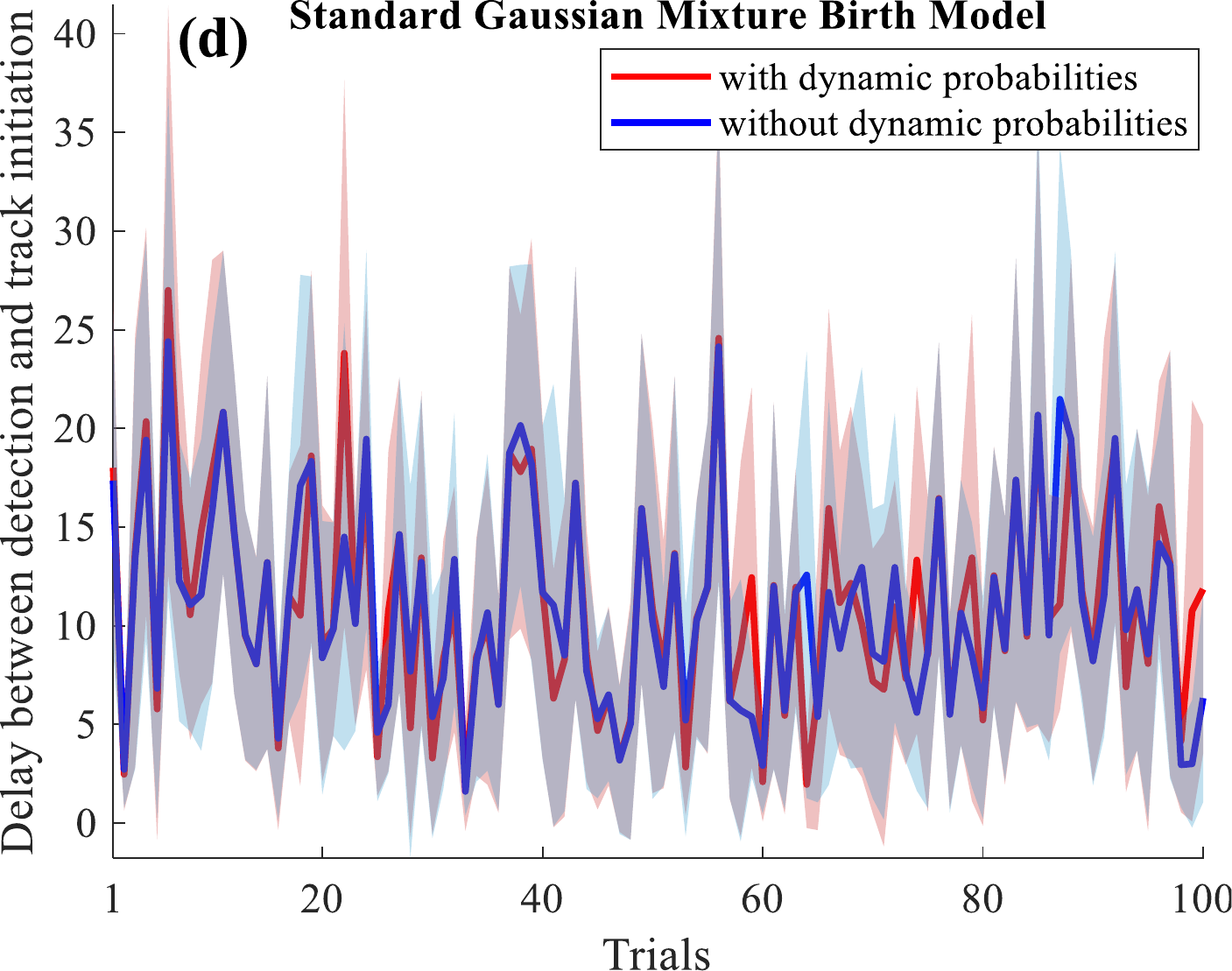}
    \includegraphics[width=0.24\linewidth]{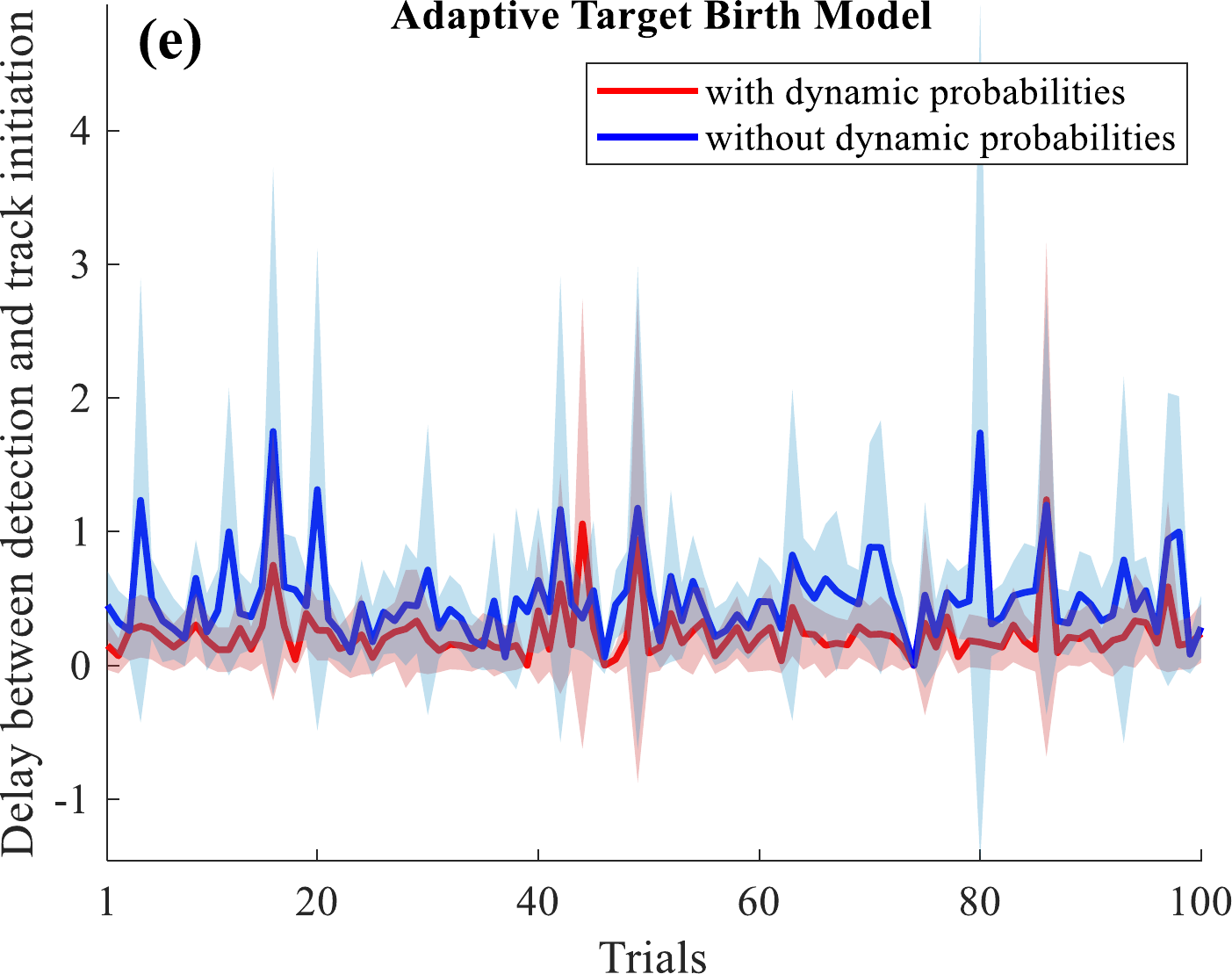}
    \includegraphics[width=0.24\linewidth]{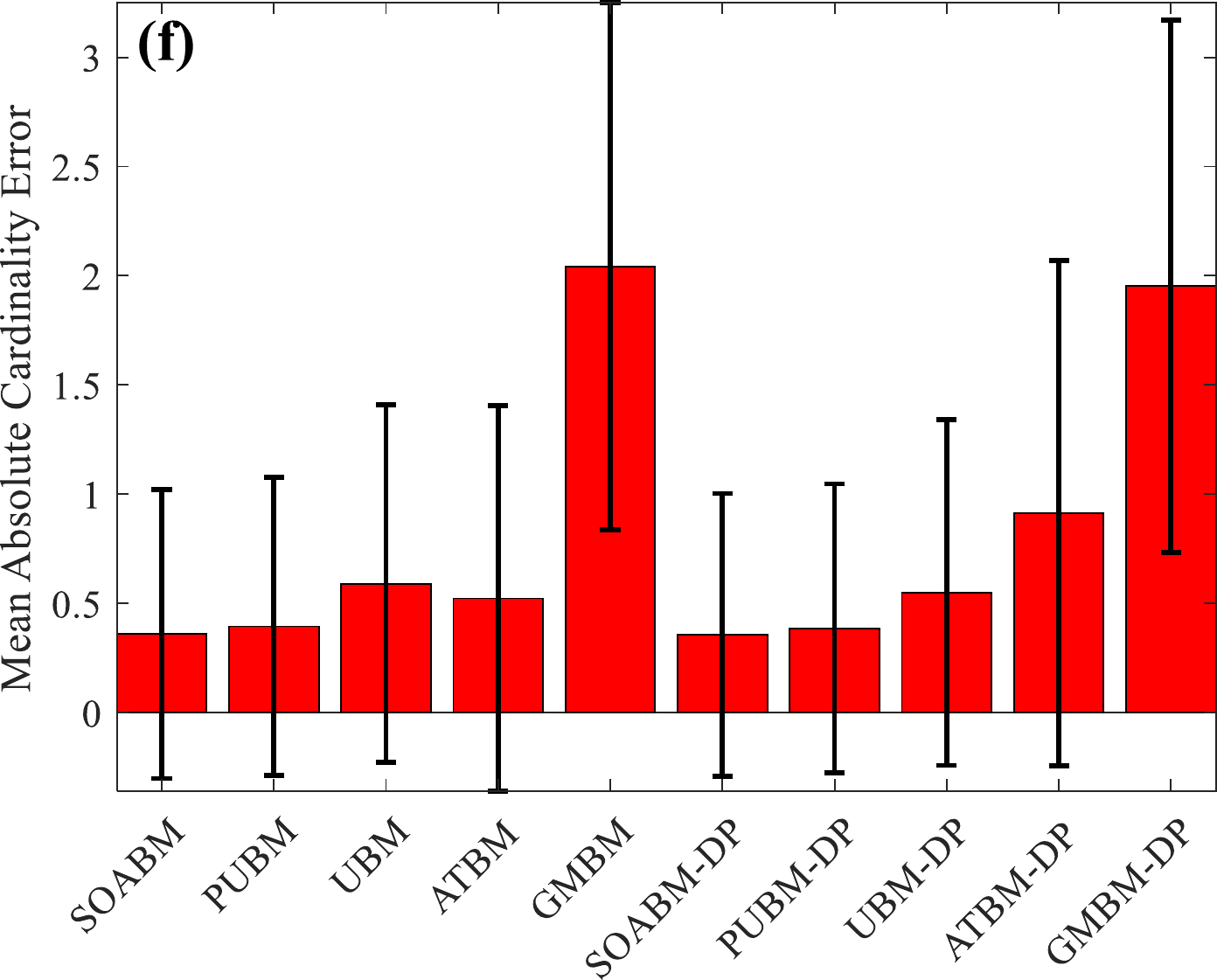}
    \includegraphics[width=0.24\linewidth]{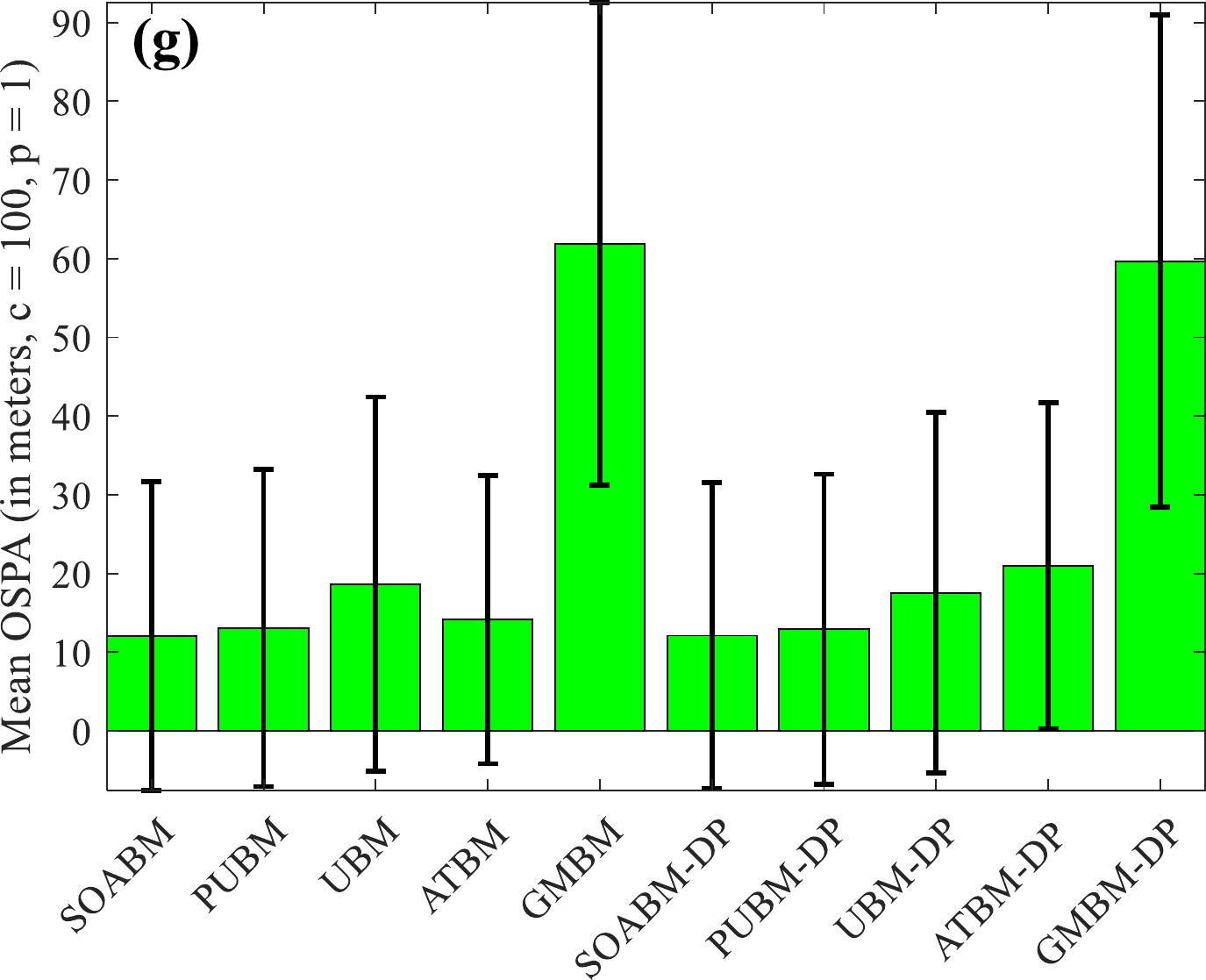}
    \includegraphics[width=0.24\linewidth]{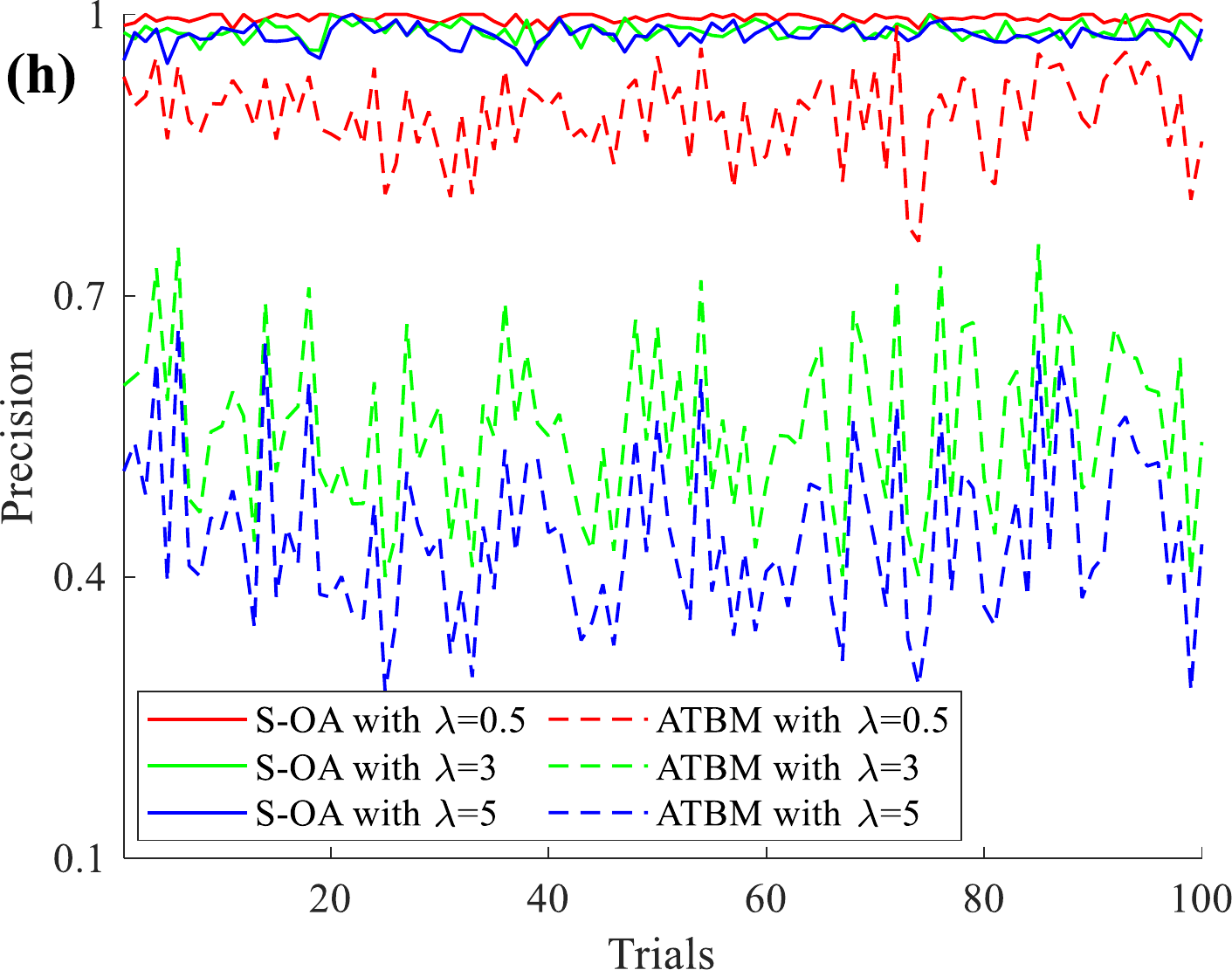}
    \caption{Subplots \textbf{(a)–(e)} illustrate the average tracking delay $\bar{k}_{\text{delay}}$ (in seconds) with $\pm0.5\sigma_{{k}_{\text{delay}}}$ bounds over 100 trials, comparing performance with and without dynamic probabilities for: \textbf{(a)} S-OA Birth Model, \textbf{(b)} Partially Uniform Birth Model, \textbf{(c)} Uniform Birth Model, \textbf{(d)} Standard GM Birth Model, and \textbf{(e)} Adaptive Target Birth Model. \textbf{(f)} Mean Absolute Cardinality Error and \textbf{(g)} Mean OSPA metric, each shown with $\pm \sigma$ over 100 trials for all birth models. \textbf{(h)} Precision curves for the S-OA and Adaptive Target Birth Models under high ($\lambda = 5$), medium ($\lambda = 3$), and low ($\lambda = 0.5$) clutter conditions.}
    \label{fig:05}
\end{figure*}

\subsection{Performance Metrics}
\label{SubSection:Performance Metrics}

To quantitatively assess performance, we define a tracking delay metric as the time difference between a target’s first detection and the moment the filter initiates its track.
For target $i$, let $k_0^{i}$ denote when it enters the scene (e.g., enters the sensor’s FOV or reappears from occlusion), $k_d^{i}$ the first detection time, and $k_t^{i}$ the track initiation time.
The delay is given by $k_{\text{delay}}^{i} = k_t^{i} - k_d^{i}$, which we aim to minimize.
A smaller delay reduces the time a target remains untracked, improving estimation accuracy and supporting safer downstream planning.
Filter performance is summarized by the mean delay $\bar{k}_{\text{delay}}$ across all targets and its standard deviation $\sigma_{{k}_{\text{delay}}}$.
We also report standard RFS-based tracking metrics.
We compute the mean of the absolute difference between the estimated and true number of targets over all time steps, along with the mean of the OSPA metric between the estimated tracks and the ground truth across all time steps.
The OSPA metric captures both localization and cardinality errors.
We use a cutoff parameter $c = 100$ and order parameter $p = 1$.

\subsection{Results}
\label{SubSection:Results}
We compare our approach with four baselines: (1) Partially Uniform Birth Model \cite{b13}, (2) Uniform Birth Model, (3) Adaptive Target Birth Model \cite{b14}, and (4) Standard GM Birth Model with fixed, predefined Gaussian birth components \cite{b12}.
For all methods, including our S-OA birth model, we implement the PHD variant from \cite{b10} that jointly estimates detection and survival probabilities.
Each filter is evaluated over 100 Monte Carlo trials of 100 seconds with randomized initial target states.
Targets follow distinct trajectories, often experiencing occlusions and path intersections.
Fig. \ref{fig:04} shows results from one trial using our S-OA GM-PHD filter.
Subplot \textbf{(a)} presents a BEV demonstrating reliable tracking without false positives from clutter, while \textbf{(b)} displays the environment overlaid with the intensity function $\nu_k(x_k)$, visualizing the Gaussian components of $\gamma_k(x_k)$ in the $XY$-plane.
Tracked targets are shown in green.


Fig. \ref{fig:05}, subplots \textbf{(a)–(e)}, present Monte Carlo simulation results for several GM-PHD filter formulations.
Using the average tracking delay $\bar{k}_{\text{delay}}$ per trial, the proposed S-OA birth model, without dynamic probabilities, outperforms the partially uniform birth model in 72 of 100 trials, the uniform and standard GM birth models in all 100 trials, and the adaptive target birth model in 5 trials.
With dynamic probabilities, it exceeds the partially uniform birth model in 68 trials, the uniform birth model in 96 trials, the standard GM birth model in all 100 trials, and the adaptive target birth model in 2 trials.
Analysis of the mean standard deviation across trials ($\bar{\sigma}_{{k}_{\text{delay}}}$) shows a similar trend.
Without dynamic probabilities, the S-OA birth model yields a mean standard deviation of 1.65 s, compared with 1.94 s (partially uniform), 2.93 s (uniform), 14.51 s (standard GM), and 0.96 s (adaptive target).
With dynamic probabilities, the corresponding values are 1.71 s, 1.82 s, 2.98 s, 14.82 s, and 0.56 s, respectively.


Although the adaptive target birth model achieves superior performance, subplot \textbf{(h)} illustrates that its precision deteriorates sharply as clutter increases ($\lambda = 0.5$ to $\lambda = 5$), due to excessive false positives.
This makes it unsuitable for applications such as autonomous driving.
In contrast, the S-OA approach maintains consistently high precision even under heavy clutter by effectively avoiding false positives.
Subplot \textbf{(f)} shows the mean absolute cardinality error across all time steps of all trials with $\pm \sigma$ bars for each method.
Without dynamic probabilities, the S-OA birth model (SOABM) reduces the error by 9.03$\%$ compared to the partially uniform model (PUBM), 39.12$\%$ to the uniform model (UBM), 31.27$\%$ to the adaptive birth model (ATBM), and 82.41$\%$ to the standard GM birth model (GMBM).
With dynamic probabilities, the reductions are 7.75$\%$, 35.18$\%$, 60.94$\%$, and 81.71$\%$, respectively.
We observe a similar trend in the mean OSPA metric across all time steps of all trials as seen in subplot \textbf{(g)}.
Without dynamic probabilities, the S-OA birth model reduces the error by 7.88$\%$ compared to the partially uniform model, 35.43$\%$ to the uniform model, 14.86$\%$ to the adaptive birth model, and 80.54$\%$ to the standard GM birth model.
With dynamic probabilities, the reductions are 6.15$\%$, 31.03$\%$, 42.4$\%$, and 79.72$\%$, respectively.
These simulation results demonstrate that the S-OA birth model outperforms the baseline methods.

\section{Experiments}
\label{Section:Experiments}
\subsection{Real-World dataset}
\label{SubSection:Real-World dataset}
We evaluate the proposed S-OA GM-PHD filter on experimental data and tracks from the KITTI dataset \cite{b20}.
The filter is used to track multiple common object classes, such as pedestrians, cars, and cyclists.
LiDAR point clouds provide the measurement set $\mathbf{Z}_k$ and semantic scene information.
We use the pre-trained 3D detector Point Voxel-RCNN \cite{b21} and the point cloud-based semantic segmentation model Cylinder3D \cite{b22}.
As the method follows a TBD framework, it is agnostic to the detector and segmentation models, allowing them to be replaced without modifying the overall framework.
A practical challenge in real-world PHD filtering is defining the birth weights $w_{\gamma,k}^{(i)}$, which depend on the expected number of new targets per time step, $\hat{N}_k$, a quantity that is difficult to estimate.
To address this, we adopt a $k$-fold cross-validation strategy.


The KITTI dataset comprises 21 sequences with varying durations and numbers of targets.
We partition the sequences into 7 bins for 7-fold cross-validation, with 3 sequences per bin.
For cross-validation, one bin is held out while the remaining 6 bins are used to estimate $\hat{N}_k$.
Specifically, for each sequence in the training folds, we compute the total number of ground-truth targets $N_{\text{seq}}$, and then obtain the value for $\hat{N}_k$ by taking a duration-weighted average across all sequences in the dataset.
The estimated value of $\hat{N}_k$ is used to define the birth weights $w_{\gamma,k}^{(i)}$ (see Section \ref{SubSection:Combined Birth Intensity}) of the Gaussian components for the filter applied to the held-out bin.
Equal weights are assigned to all birth components.
The procedure is repeated for all 7 folds, yielding a complete evaluation of filter performance on the KITTI validation dataset.
We compare the S-OA birth model against the partially uniform and the uniform birth model.

\begin{figure}
    \centering
    \includegraphics[width=0.75\linewidth]{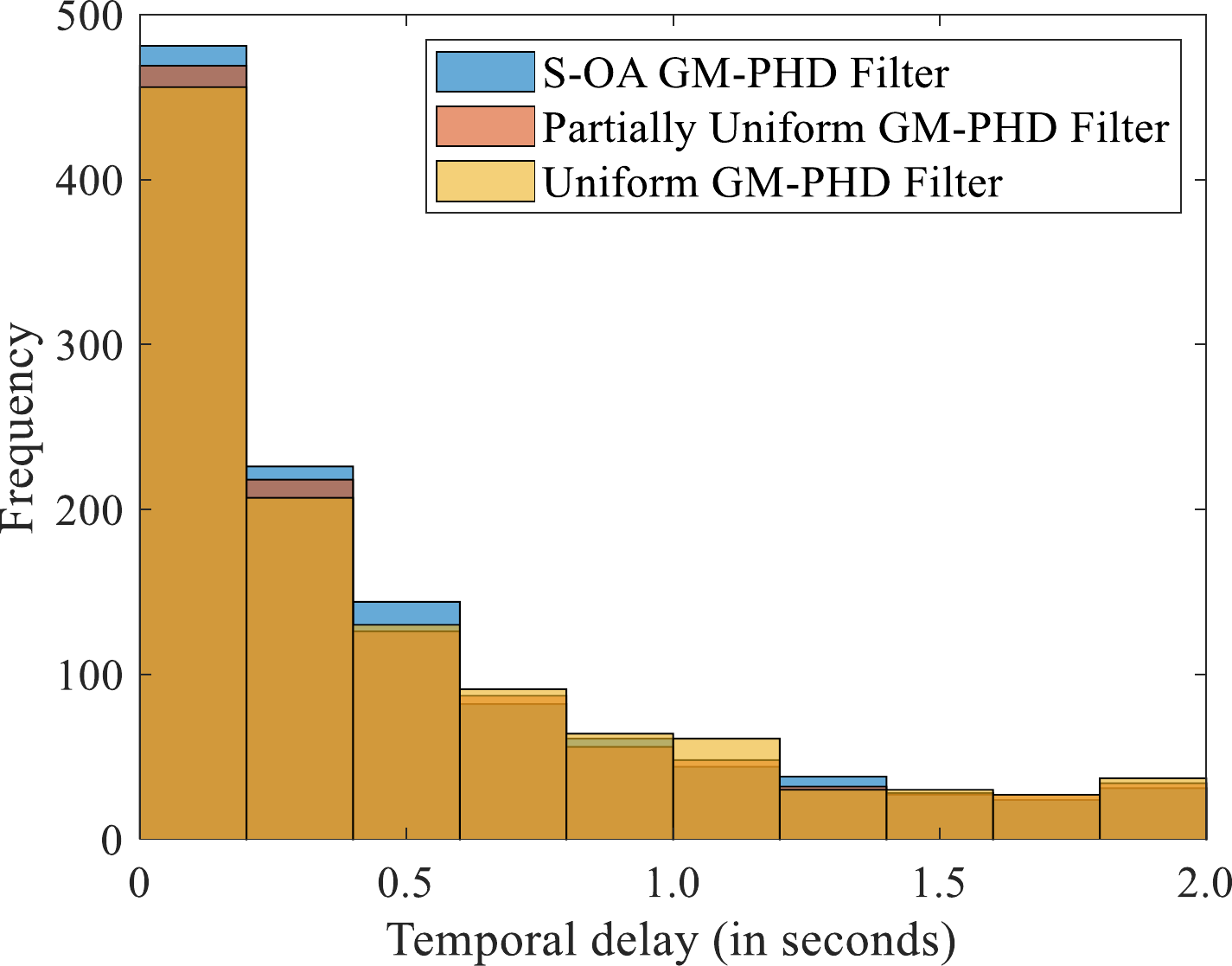}
    \caption{Histogram of tracking delays $k_{\text{delay}}^i$ (in seconds) for all targets in the KITTI dataset, comparing the three birth model formulations.}
    \label{fig:06}
\end{figure}

\begin{figure}
    \centering
    \includegraphics[width=0.70\linewidth]{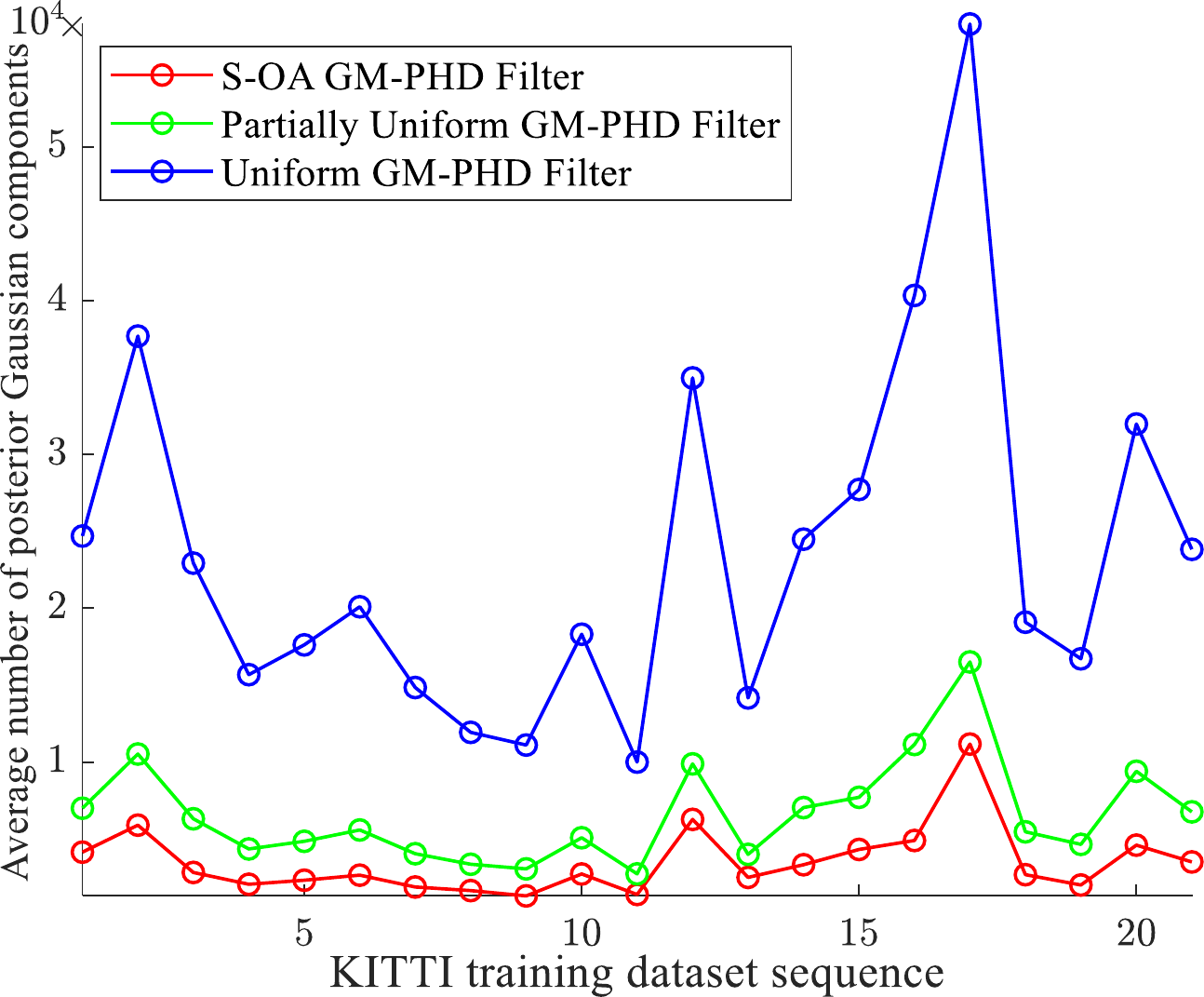}
    \caption{Average number of components in the GM of the posterior intensity for each sequence in the KITTI tracking dataset, evaluated across all models.}
    \label{fig:07}
\end{figure}

\begin{table}
\caption{Mean Absolute Cardinality Error and Mean OSPA Metric over all time steps on the KITTI dataset for each method.}
\centering
\renewcommand{\arraystretch}{1.2}
\begin{tabularx}{\columnwidth}{|p{2cm}|X|p{1.7cm}|X|}
\hline
{\textbf{Filter}} 
& \textbf{S-OA GM-PHD Filter} (ours) 
& \textbf{Partially Uniform GM-PHD Filter} 
& \textbf{Uniform GM-PHD Filter} \\ 
\hline
Mean Absolute Cardinality Error & $\mathbf{6.5565~\pm}$ $\mathbf{4.4513}$ & ${6.7194~\pm}$ ${4.5109}$ & ${6.7836~\pm}$ ${4.4926}$ \\ 
\hline
Mean OSPA Metric (meters) & $\mathbf{7.8666~ \pm}$ $\mathbf{1.9936}$ & ${8.0362~\pm}$ ${1.9943}$ & ${8.1023~\pm}$ ${1.9742}$ \\ 
\hline
\end{tabularx}
\label{Table:01}
\end{table}

To evaluate the performance of the three methods, we compute the tracking delay metric $k_{\text{delay}}^i$ for each target $i$ in the KITTI dataset for each method.
Using pairwise comparison across the three methods, we find that our S-OA birth model matches or outperforms the partially uniform birth model in 70.3\% of cases, and the uniform birth model in 72.2\% of cases.
Fig. \ref{fig:06} presents a histogram of the delay values $k_{\text{delay}}^i$ across all targets for the three methods.
The distribution shows that our method achieves the highest frequency in the smallest delay bins, whereas the baseline approaches dominate in the larger delay bins.
This indicates that, in the majority of cases, our method detects targets earlier than the alternative approaches.
Table \ref{Table:01} reports the mean absolute cardinality error and the mean OSPA metric (with $c = 10$ and $p = 1$) across all time steps, including their corresponding $\pm \sigma$ values for each method.
For both metrics, the proposed approach demonstrates improved performance.

We further assess computational cost on the KITTI dataset by examining the number of Gaussian components in the posterior intensity of the filter.
A larger number of components increases cycle time and reduces individual component weights, potentially delaying target birth and track initiation.
Fig. \ref{fig:07} shows the average number of posterior components per filter across all sequences.
The uniform birth model yields the most components, followed by the partially uniform model and our S-OA birth model.
Notably, the uniform model produces about 5–6 times more components than ours, many of which contribute little to tracking.
In contrast, our method is significantly more computationally efficient, making it better suited for real-time deployment.


\subsection{Parameter Sensitivity Analysis}
\label{SubSection:Parameter Sensitivity Analysis}
\begin{figure}
    \centering
    \includegraphics[width=0.85\linewidth]{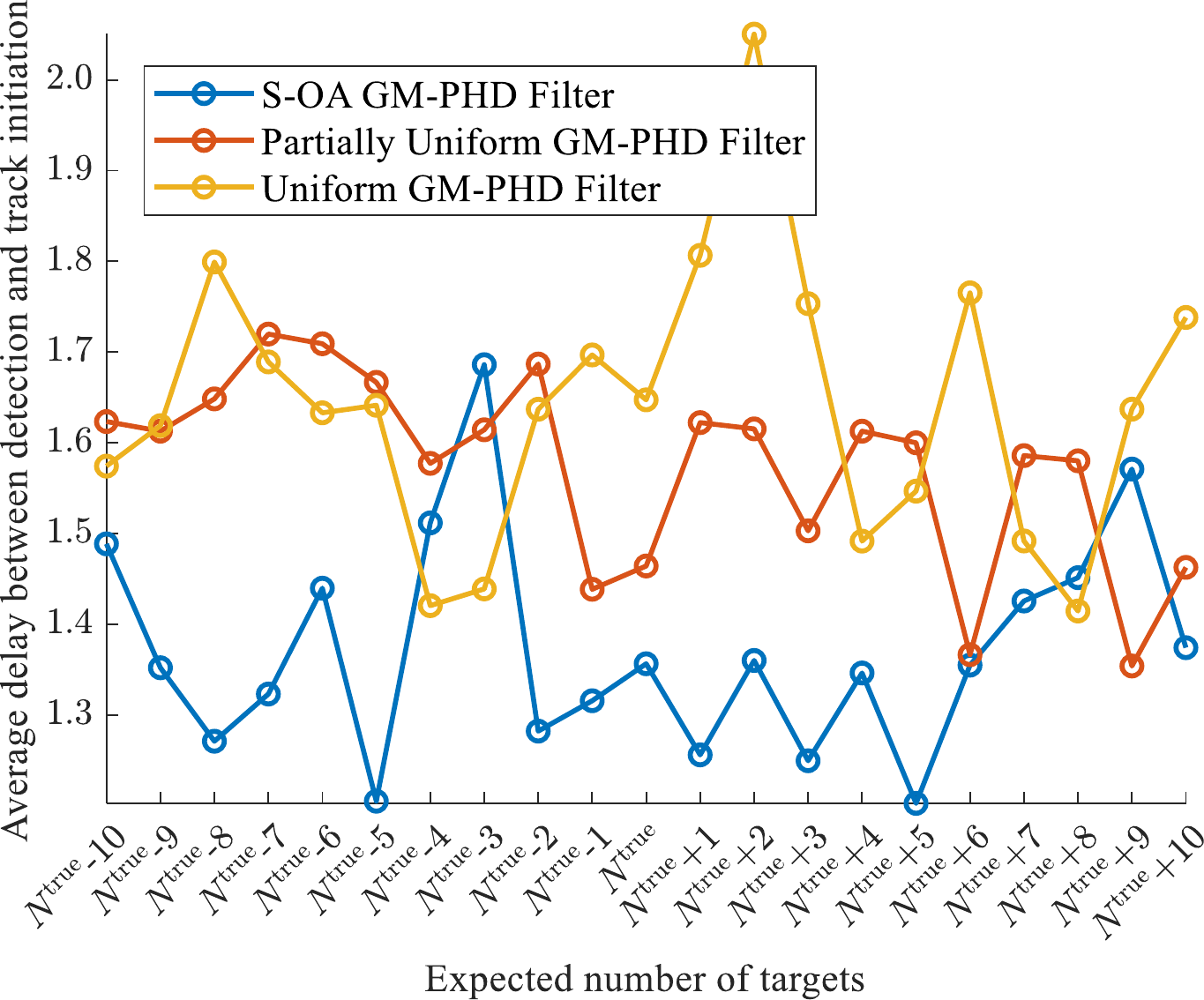}
    \caption{Average delay plot (in seconds) for the three methods on the KITTI dataset with different values of $\hat{N}_k$.}
    \label{fig:08}
\end{figure}

As stated previously, a major challenge in deploying PHD filters for real-time operations is determining the birth weight $w_{\gamma,k}^{(i)}$.
Most prior research only focuses on simulation studies and not experimental data.
This leads to the question of how sensitive the approaches are to parameter tuning in experiments.
The birth model designed for the filter should be able to accommodate these variations to the extent possible.
In section \ref{SubSection:Real-World dataset} we presented an approach to empirically estimate birth weights using data and $k$-fold cross validation.
To determine how our birth model performed as compared to the other birth models, we tested the performance of the filters across different values of $\hat{N}_k$ (and therefore scaled values of $w_{\gamma,k}^{(i)}$).
We first determined the true value for the expected number of target births ($N^{\text{true}}$) using the ground truth data for the KITTI dataset.
We then defined 21 different values for $\hat{N}_k$ ranging from $N^{\text{true}} - 10$ to $N^{\text{true}} + 10$, and ran the GM-PHD filter with the partially uniform, uniform, and our S-OA birth model.
These trials yield the average delay $\bar{k}_{\text{delay}}$ for the entire KITTI tracking dataset for each value of $\hat{N}_k$, which are summarized in Fig. \ref{fig:08}.
The results indicate that, in most scenarios, the S-OA birth model achieves a lower average tracking delay compared to both the partially uniform and uniform birth models.
Additionally, the S-OA model demonstrates reduced performance variability across different weight parameter settings, with a standard deviation of 0.12 s for the plotted means, which is lower than that of the uniform birth model (0.15 s), though slightly higher than the partially uniform birth model (0.10 s).
These findings suggest that the S-OA birth model not only delivers improved performance but also exhibits greater robustness to parameter variations, an important advantage for real-world applications where extensive parameter tuning may not be feasible.

\section{Conclusion}
\label{Section:Conclusion}
In this work, we introduced the S-OA birth model, a novel GM-PHD filter birth formulation that explicitly integrates occlusion reasoning and scene semantics within the RFS framework.
Unlike conventional models, it adaptively places birth components along occlusion boundaries, within semantically meaningful regions, and near the sensor’s FOV limits.
Whereas many learning-based methods delay track initiation until a target remains unassociated for several time steps, our approach leverages scene context to initiate tracks earlier while limiting false positives.
By hypothesizing new targets in likely emergence areas, the method improves robustness in cluttered and dynamic environments.
Extensive Monte Carlo simulations show consistently lower tracking delays and reduced variability compared to baselines.
Experiments on the KITTI dataset further confirm earlier target initialization in most cases relative to uniform and partially uniform models.
Sensitivity analysis demonstrates that the S-OA model is less affected by birth weight parameter variations, underscoring its robustness in practical settings where tuning is difficult.
Overall, incorporating semantic priors and occlusion awareness into RFS filtering yields a more reliable and effective birth process for multi-target tracking in autonomous driving.

\end{document}